\documentclass[10pt,twocolumn,letterpaper]{article}

\usepackage{cvpr}              %

\usepackage{graphicx}
\usepackage{amsmath}
\usepackage{amssymb}
\usepackage{booktabs}
\usepackage{cprotect}
\usepackage{pifont}
\newcommand{\cmark}{\ding{51}}%
\newcommand{\xmark}{\ding{55}}%

\usepackage[accsupp]{axessibility}  %

\usepackage[pagebackref,breaklinks,colorlinks]{hyperref}

\usepackage[capitalize]{cleveref}
\crefname{section}{Sec.}{Secs.}
\Crefname{section}{Section}{Sections}
\Crefname{table}{Table}{Tables}
\crefname{table}{Tab.}{Tabs.}

\def\cvprPaperID{10287} %
\def\confName{CVPR}
\def\confYear{2023}

\begin{document}

\title{Language-Guided Music Recommendation for Video via Prompt Analogies}

\author{Daniel McKee$^{1}$\thanks{Work done as an intern with Adobe Research}\hspace{8mm}Justin Salamon$^2$\hspace{8mm}Josef Sivic$^{2,3}$\hspace{8mm}Bryan Russell$^2$ \vspace{0.3em}\\
$^1$University of Illinois at Urbana-Champaign\hspace{4mm}$^2$Adobe Research \\
$^3$Czech Institute of Informatics, Robotics and Cybernetics, Czech Technical University
\\
{\tt\small dbmckee2@illinois.edu
\hspace{3mm}
salamon@adobe.com
\hspace{3mm}
josef.sivic@cvut.cz
\hspace{3mm}
brussell@adobe.com
} \\
\small{\url{https://www.danielbmckee.com/language-guided-music-for-video}}
}
\maketitle

\begin{abstract}

We propose a method to recommend music for an input video while allowing a user to guide music selection with free-form natural language. A key challenge of this problem setting is that existing music video datasets provide the needed (video, music) training pairs, but lack text descriptions of the music. This work addresses this challenge with the following three contributions. First, we propose a text-synthesis approach that relies on an analogy-based prompting procedure to generate natural language music descriptions from a large-scale language model (BLOOM-176B) given pre-trained music tagger outputs and a small number of human text descriptions. Second, we use these synthesized music descriptions to train a new trimodal model, which fuses text and video input representations to query music samples. For training, we introduce a text dropout regularization mechanism which we show is critical to model performance. Our model design allows for the retrieved music audio to agree with the two input modalities by matching visual style depicted in the video and musical genre, mood, or instrumentation described in the natural language query. Third, to evaluate our approach, we collect a testing dataset for our problem by annotating a subset of 4k clips from the YT8M-MusicVideo dataset with natural language music descriptions which we make publicly available. We show that our approach can match or exceed the performance of prior methods on video-to-music retrieval while significantly improving retrieval accuracy when using text guidance.
\end{abstract}

\begin{figure}
\begin{center}
\includegraphics[width=0.475\textwidth]{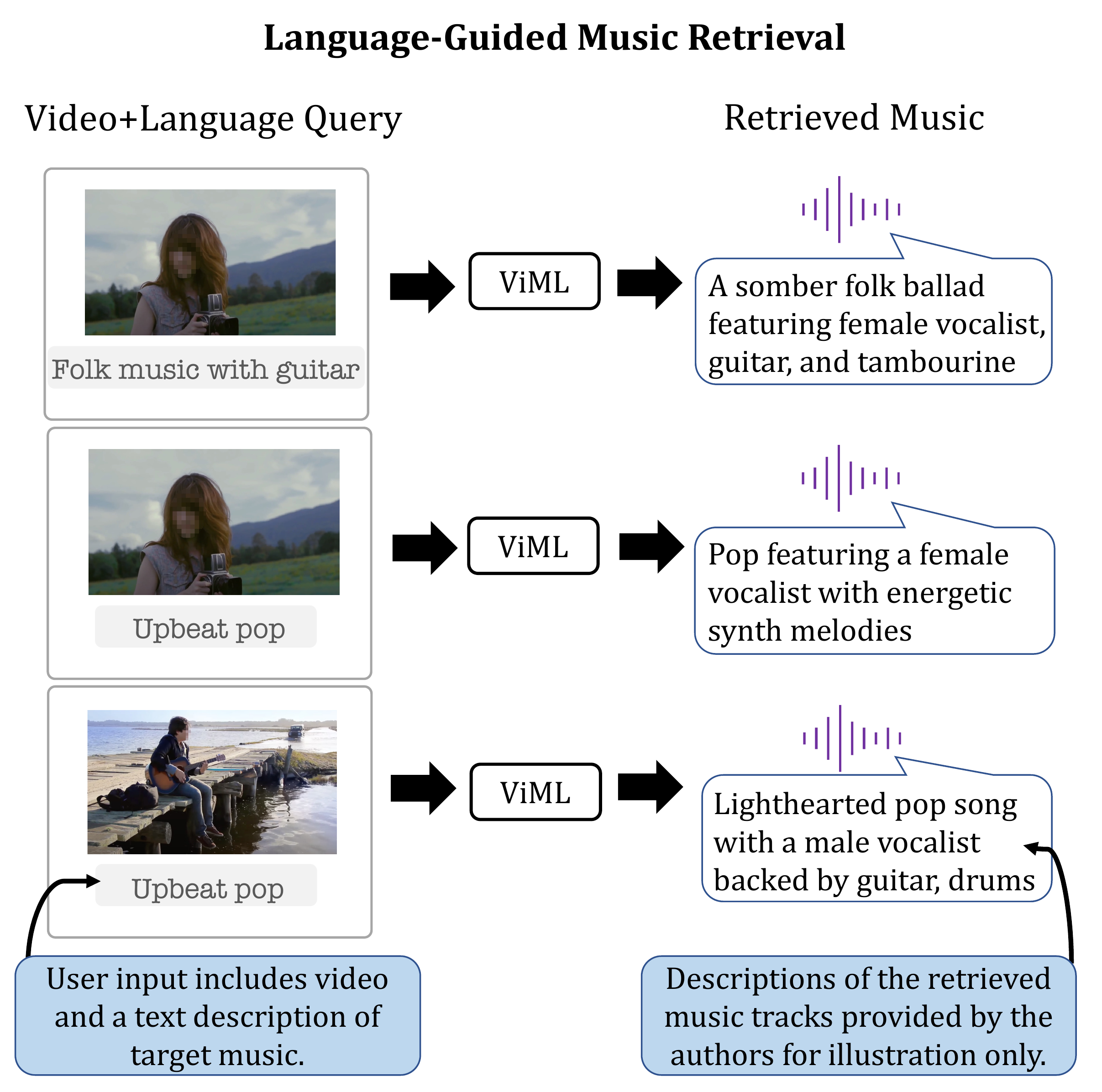}
\caption{
{\bf Language-guided music retrieval.} 
Our ViML model takes a video and text prompt as input to retrieve a suitable music track from a database. The model learns to fuse video and language representations in order to guide retrieval.
Notice how our approach retrieves audio matching both the video and language content. For the same video query (top two rows), we can change the music style to match the language query, and for the same \texttt{Upbeat pop} query (bottom two rows) we can change the vocalist to match the video content. {\bf To fully appreciate our results, please view and listen to the companion video on our website}. 
}
\label{fig:teaser}
\end{center}
\end{figure}

\section{Introduction}

A key part of the video editing process for creators is choosing a musical soundtrack. Especially given the rise of short-form videos on social media platforms, automated music recommendation systems have become an increasingly common and important part of video editing applications. While these systems can be helpful for finding relevant music, they often provide limited capability for user control over the types of music recommended.
In previous work, music is retrieved based solely on the visual content and style from a video \cite{suris2022s,pretet2021cross}. However, music itself can convey critical information about how a video should be perceived. Music selection alone can transform a visual scene into one that is perceived as happy, scary, or sad\footnote{\url{https://www.youtube.com/watch?v=iSkJFs7myn0}}. As a result, the lack of user input capability to describe a target music for an inputted video is a key limitation on the utility of current music recommendation methods.

In this work, we propose a more flexible music-for-video recommendation approach that allows a user to guide recommendations towards specific musical attributes including mood, genre, or instrumentation, illustrated in Figure~\ref{fig:teaser}. To maximize flexibility and user convenience, we propose to take user musical attribute descriptions in the form of \textit{free-form natural language} (\eg, ``Folk music with guitar" in Figure \ref{fig:teaser}). There are two key challenges in learning a model for language-guided music recommendation for video.  
First, while there are datasets which include music+text \cite{law2009evaluation,bertin2011million,oramas2017multi,bogdanov2019mtg} or music+video \cite{abu2016youtube}, there are no available datasets which include music, video, and text together. Further, the existing datasets that do include text and music focus on a limited vocabulary of tags rather than free-form text.
Second, previous works have explored jointly learning visual, audio, and text embeddings \cite{aytar2017see,alayrac2020self,zellers2022merlot,rouditchenko2020avlnet,akbari2021vatt}, and without careful
regularization, a network can overfit and possibly learn to
ignore one of the input modalities. We seek to train a model
that keeps the information flow through the network and
does not ignore one of the modalities.

\begin{figure*}
  \begin{center}
\includegraphics[width=1.0\textwidth]{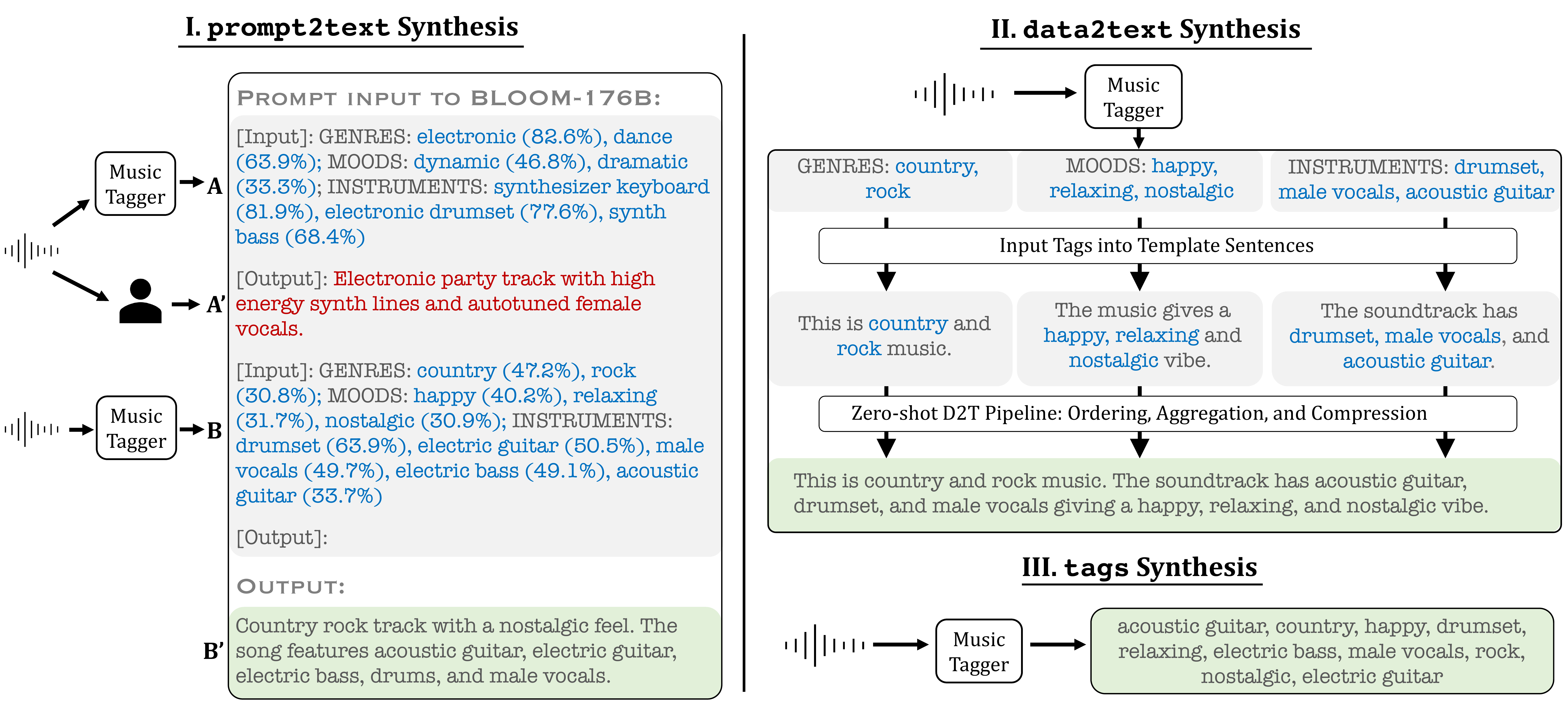}
  \caption{
\textbf{Overview of three text synthesis approaches explored in our work.} All rely on tag predictions from a pretrained music tagger model. We highlight output text from each method in green, inputs from the tagger in blue font, and inputs from a human annotator in red font.
\textbf{Left:} We introduce the \texttt{prompt2tags} approach for generating natural language descriptions given automatically predicted music tags and a small set of human descriptions. We ask a large language model (BLOOM-176B) to complete an analogy task ($A:A^\prime :: B:B^\prime$) between music tags ($A, B$) and descriptions ($A^\prime, B^\prime$). 
\textbf{Top right:} The \texttt{data2text} pipeline inserts sampled tags into randomly selected template sentences corresponding to each tag category.
The Zero-shot D2T model \cite{kasner2022neural} then orders, aggregates, and compresses these templates into a final output description.
\textbf{Bottom right:} The \texttt{tags} approach involves direct concatenation of high confidence tags to form the text description of the music.
  }
\label{fig:text_synthesis}
  \end{center}
  \end{figure*}

In order to meet the challenges outlined above,  our work makes the following contributions:

(1) We propose a new approach to automatically generate natural language descriptions for a music video dataset. This approach combines a pre-trained music tagger with a large-scale language model to output natural language descriptions for any music clip, illustrated in Figure~\ref{fig:text_synthesis} (left). First, the tagger predicts tags from a pre-defined vocabulary describing musical genre, mood, or instrumentation. Second, these predicted tags, together with their probabilities, are converted 
into a rich natural language description for the music video using a carefully designed large-scale language model prompting procedure based on analogies with a small number of human-provided text descriptions (\ie, $A$ (\texttt{tags}) : $A^\prime$ (\texttt{description}) :: $B$ (\texttt{tags}) : $B^\prime$ (\texttt{description})), where  $A$ and $B$ are music tags automatically provided by the tagger, $A^\prime$ is a human-provided text description, and $B^\prime$ is the natural language description output by the large-scale language model.

(2) We propose a Transformer-based model architecture with a video-text fusion module.
Our model, which we call \textbf{Vi}deo to \textbf{M}usic with \textbf{L}anguage (ViML), is able
to retrieve music that matches both the visual content/style of the input video and described musical genre, mood, and instrumentation in the natural language query. 
Similar to prior work \cite{hussen2020modality,neverova2015moddrop,wang2020makes}, we find that training
with text dropout as a regularization mechanism is critical to achieve music retrieval performance improvements from added text inputs.

(3) We release a dataset of 4000 high quality text annotations for clips from a subset of the YT8M-MusicVideo dataset \cite{abu2016youtube} to evaluate language-guided music recommendation. We show that our method can achieve substantial improvements over prior works on music retrieval when incorporating text inputs. Moreover, our model can match or even exceed performance of baseline music-for-video recommendation models when the text input is ignored.

\section{Related Work}

\noindent \textbf{Music and language.} There are numerous music tagging datasets which contain tags specifying attributes like mood, genre, or instrumentation \cite{law2009evaluation,bertin2011million,oramas2017multi,bogdanov2019mtg}, and several works have studied training automated music taggers from such datasets \cite{pons2019musicnn,choi2016automatic,won2020eval,lee2020metric,lee2017sample,won2021transformer}. Beyond these methods constrained to limited tag vocabularies, some works also have studied jointly embedding music and free-form natural language \cite{won2020eval,choi2019zero,huang2022mulan,manco2022contrastive}. However, none of these approaches incorporate the video modality.

\vspace{0.5em}
\noindent \textbf{Music recommendation for video.} Others have investigated automatic recommendation of music based on style and content of an input video \cite{hong2017content,zeng2018audio,li2019query,pretet2021cross,suris2022s}. Pr\'etet et al. \cite{pretet2021cross} build on previous self-supervised methods \cite{hong2017content} by incorporating learned audio features instead of handcrafted features. More recently, Sur\'is et al. \cite{suris2022s} propose the MVPt model which employs a self-supervised contrastive loss and Transformer \cite{vaswani2017attention} architecture to greatly improve the long-range temporal context modeling in order to retrieve suitable music for a given input video. However, none of these approaches incorporate the natural language modality which we focus on in this work.

\vspace{0.5em}
\noindent \textbf{Video, audio, \& language.} While a wide variety of works have explored audio-visual or vision-language topics, a smaller number focus on jointly embedding video, audio, and language \cite{aytar2017see,zellers2022merlot,akbari2021vatt,rouditchenko2020avlnet,guzhov2022audioclip,wu2022wav2clip}.
Specifically, Alayrac et al. \cite{alayrac2020self} investigate how best to combine audio and video with text representations.
The VATT model \cite{akbari2021vatt} is a fully end-to-end tri-modal model capable of using a single shared Transformer backbone across modalities. 
Lastly, two recent methods \cite{guzhov2022audioclip,wu2022wav2clip} extend CLIP \cite{radford2021learning} to jointly embed audio. While relevant, all of these approaches share a common focus on ``environmental" or ``everyday" sounds rather than music, and they lack the long-range temporal context modeling critical for music recommendation as a result. In addition, none of these works address a downstream problem of using two modalities in combination (video, text) to query results from another (music).

\vspace{0.5em}
\noindent \textbf{Few-shot language model prompting.} Recent large language models have shown significant success at a wide variety of few-shot or zero-shot tasks 
from those related to reading comprehension and QA \cite{brown2020language,wei2021finetuned}, to reasoning \cite{wei2022chain,kojima2022large}, or even
data augmentation \cite{ding-etal-2020-daga,wang-etal-2022-promda,yang-etal-2020-generative}. A few works have extended this success to multimodal applications. For example, Zeng et al. \cite{zeng2022socratic} show that language models can solve video understanding or image captioning tasks by reformulating these problems as reading comprehension or QA tasks with inputs from large visual or audio models. Other works have used language models to help with generating or retrieving text annotations for multimodal tasks \cite{lin2022learning} or in robotic planning \cite{huang2022inner,singh2022progprompt}.
In this work, we propose a completely new application of few-shot language query modeling: generating free-form musical text descriptions from music tags.

\section{Approach}

\begin{figure}
  \begin{center}
  \includegraphics[width=0.5\textwidth]{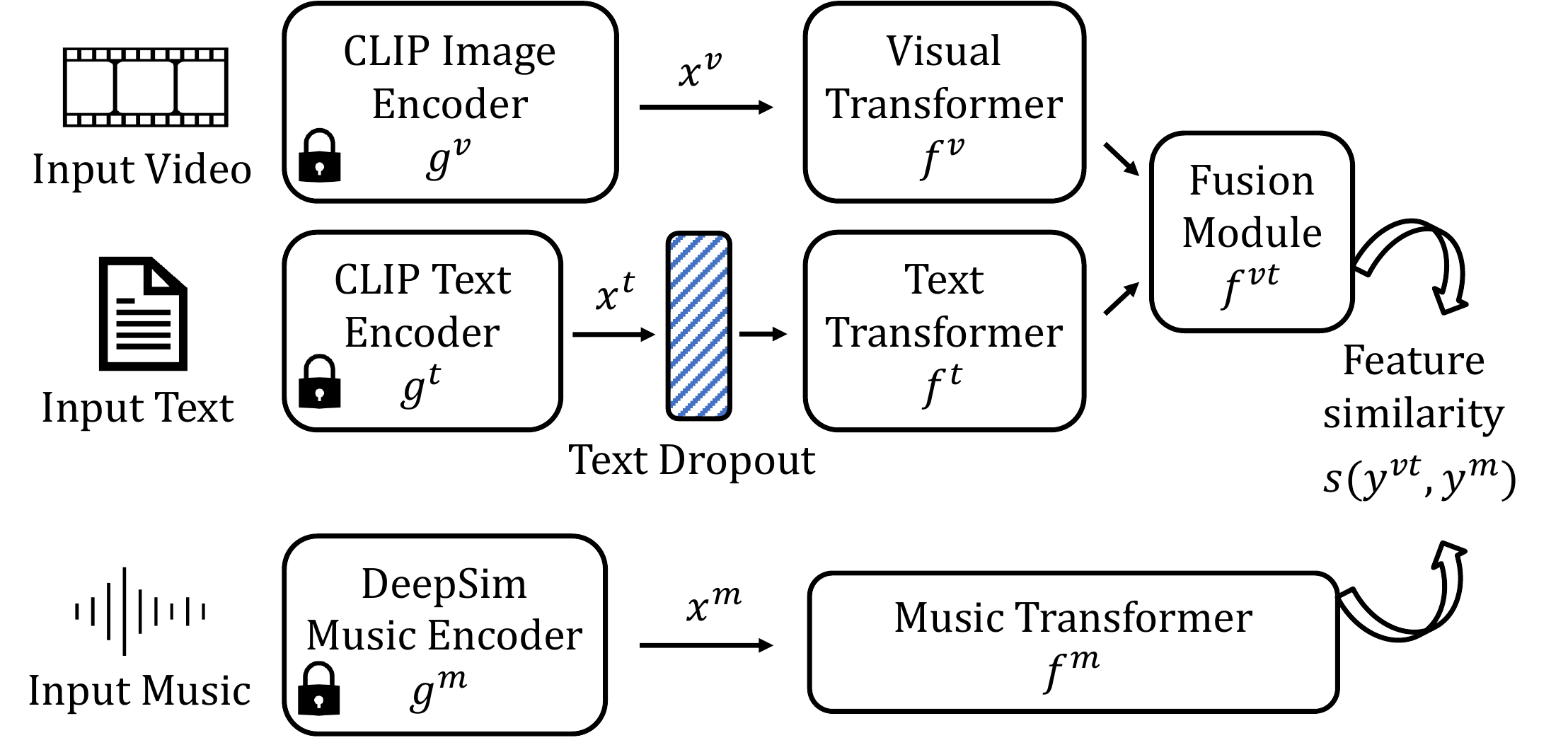}
  \caption{Our proposed ViML model embeds inputs from three modalities (video, text, and audio) into an embedding space. We extract base features using DeepSim \cite{lee2020metric} for music input and from CLIP \cite{radford2021learning} for video frames and text descriptions. These base features are inputted to Transformer encoders for each modality. The video and text features are combined with a fusion module to enable querying of music in a shared embedding space. Finally, we employ text dropout to address the difference in granularity between the three modalities. Since video is a more complex input modality, text dropout forces an improved video representation by preventing co-adaptation of the video and text representations.
  }
  \label{fig:model_diagram}
  \end{center}
  \end{figure}

Our goal is to train a pair of feature encoders $f^{vt}$ and $f^{m}$ which are capable of predicting the similarity $s(f^{vt},f^{m})$ between an input pair of video and musical text description  $(v, t)$ and a music clip $m$, as illustrated in Figure~\ref{fig:model_diagram}. To train such a model in a supervised manner, it is necessary to have a dataset of corresponding triplets $(v, m, t)$. While large-scale datasets of videos with paired music are available, it is difficult to find datasets which also contain high-quality natural language descriptions of the paired music tracks. As a result, we investigate a synthesis approach based on a model $G$ which generates text descriptions from available structured data in the form of music tags for each music track. In the following sections, we first discuss the musical description synthesis approach $G$ before describing an approach to train a language-guided video-to-music recommendation model.

\subsection{Synthesizing Text Descriptions for Music}
\label{sec:synthesis}

Suppose that we are given a set of video and music audio pairs $(v_i, m_i)$ and that we also have access to structured data $d_i \in \mathcal{T}^D$ which describe the music $m_i$. 
In our case, this structured data consists of musical tags with confidences. 
Each music track $m_i$ may be described by a free-form human text description $t_i \in \mathcal{T}^T$. However, it can be prohibitively expensive to obtain high-quality human descriptions on a large scale. Instead, we propose to synthesize such text descriptions using a generator $G:\mathcal{T}^D\rightarrow \mathcal{T}^T$ which maps structured data describing an audio track to the space of natural human descriptions. 

The goals of such a mapping function $G$ are that: 
(i) a predicted output $\Tilde{t}_i = G(d_i)$ should preserve the semantic meaning contained within the structured data $d_i$ corresponding to a specific musical track and (ii) the distribution of predicted outputs $\Tilde{t}_i$ should follow the distribution of ground truth human text annotations $t_i \in \mathcal{T}^T$. Training a fully supervised model to be the generator function $G$ would require a large quantity of human text descriptions. Instead, we explore zero-shot or few-shot approaches to obtain a generator function $G$.
In particular, we describe three approaches that all use the automatically predicted music tags: a \texttt{prompt2text} approach which relies entirely on careful few-shot prompting of a pretrained language model, 
a zero-shot \texttt{data2text} approach which rephrases templated sentences using pretrained language models,
and a zero-shot \texttt{tags} baseline that represents the music track description directly via the set of automatically obtained tags. Details are given next.

\vspace{0.5em}
\noindent \textbf{I. Few-shot \texttt{prompt2text} approach.}
We first explore whether the full mapping function $G$ can be encompassed by a single large language model through careful few-shot prompting. This approach relies on a small set of example human-provided descriptions ${t_0, ..., t_N}$ where $t_i \sim \mathcal{T}^T$. We assume that for each example $t_i$, we also have a paired structured data output $d_i$, provided by the automatic music tagger, which describes the same audio track. 
Unlike prior prompt-based data augmentation works \cite{ding-etal-2020-daga,wang-etal-2022-promda,yang-etal-2020-generative} which aim for an unconditional generator $G$, we aim to generate text data $\tilde{t}_i\sim G(d_i)$ conditioned on structured data $d_i$ such that it follows the distribution of human sentences $\tilde{t}_i\in\mathcal{T}^T$.

As shown in Figure \ref{fig:text_synthesis} (left), the structured data output $d_i$ is converted to text form via a template, and a set of pairs $(d_0, t_0), ... , (d_k, t_k)$ are used to form $k$ input/output components in the prompt. The final segment of the prompt is the structured data $d_i$ corresponding to a new music track. Given $d_i$, the model will attempt to output a description $t_i$ following the mapping $\mathcal{T}^D\rightarrow \mathcal{T}^T$ suggested by the example inputs.
For text generation in this setting, we use the BLOOM-176B \cite{bloom} model which is trained on a highly diverse 1.5TB text corpus.

Given that the \texttt{prompt2text} allows for the greatest freedom in generation, the model can more easily generate a diverse set of text resembling the target distribution $\mathcal{T}^T$.  The \texttt{prompt2text} approach is also very flexible as large language models like BLOOM can handle a variety of different structured data inputs such as both tags and their confidence predictions. However, the model may also be less likely to preserve semantic meaning from structured data.

\vspace{0.5em}
\noindent \textbf{II. Zero-shot \texttt{data2text} approach.} 
The second setting that we propose involves a data-to-text generation process which is illustrated in Figure~\ref{fig:text_synthesis} (top right). At a high level, the goal of this method is to insert structured tag data into predefined template sentences and rephrase these template sentences using a language model while preserving original semantic meaning.
We begin with the tags predicted for each music track and grouped into genre, mood, and instrument categories. 
We define a set of category-specific templates in the form of short sentences with placeholders for tags.
We randomly sample a template sentence for each category, and fill the template with the high-confidence predicted tags for that category. 
To form these sentences into more natural free-form descriptions, we make use of pretrained large language models. Specifically, we follow the Zero-shot D2T approach \cite{kasner2022neural}, which consists of ordering, aggregation, and compression modules built on pretrained RoBERTa \cite{liu2019roberta} and BART \cite{lewis2019bart} language models. The pipeline components first set the order of the individual filled template sentences and assign which sentences should be combined into a single sentence. Next, the compression module uses a generative text model to rewrite the input sentences based on the ordering and aggregation specifications. The module aims to rephrase the information while preserving semantic meaning. Because this D2T pipeline makes use of models that are pre-trained on large, general text corpuses, we find these modules to perform well at generating music descriptions in a zero-shot manner.

\vspace{0.5em}
\noindent \textbf{III. \texttt{tags} approach.} The final setting we use involves a simple concatenation of predicted tags.  We take the set of top filtered predicted tags for each music track (this set typically numbers around 10-15 tags total).
We then randomly shuffle these tags to prevent model dependence on ordering and concatenate all of the tags into a comma-separated list of musical descriptions (\eg, ``synthesizer keyboard, electronic drumset, pop, dance, synth bass, electronic, happy, electric guitar, frantic, dynamic"). While this approach strongly preserves the semantic meaning, it fails to generate text with diverse vocabulary and form which would well represent the human annotation distribution $\mathcal{T}^T$.

\subsection{Text Dropout for Music Retrieval Training}
\label{sec:method}

The objective here is to retrieve music track $m$ matching a query video $v$ and a natural language query $t$ describing the target music track. This is a challenging task as the model has to fuse together information from both the input video and the input language query to then find a semantically appropriate music track. Moreover, the difference in granularity between audio/video and text can significantly hinder training.
We design a tri-modal approach, dubbed ViML, for this task and introduce text dropout to address the granularity issue. 
In a similar manner to the way dropout prevents overfitting by reducing co-adaptation between individual neurons \cite{srivastava2014dropout}, text dropout serves to avoid overfitting to the text inputs and prevent co-adaptations between the video and text encoders.
The approach is illustrated in Figure~\ref{fig:model_diagram} and details of model architecture, loss, and text dropout are given next.

\vspace{0.5em}
\noindent \textbf{Model architecture.} Our model is trained on a set of (video, music, text) pairings, $(v, m, t)$, corresponding to a music video clip $v$, which has been labeled with a generated text description $t$ of its music track $m$, as outlined in Section~\ref{sec:synthesis}. We transform these inputs into base features $x^v = g^v(v)$ for visual video features, $x^m= g^m(m)$ for music features, and $x^t = g^t(t)$ for text features using pretrained large-scale encoders $g^v$, $g^m$, and $g^t$ which are frozen during training. 

Each base feature representation $x$ is of dimension $n \times d$, where $n$ is the length of the temporal sequence of  base features representing a video clip and $d$ is the dimension of the base feature. We note that while our model is capable of handling a sequence of temporal text descriptions similar to music or video, we obtain only a track-level text description in practice meaning that $n=1$ for text.

Our tri-modal model consists of three separate modules corresponding to each modality $f^v, f^m, f^t$, and a fourth fusion module $f^{vt}$ to combine video and text representations. The modules take respective base features and output embeddings $y^v=f^v(x^v)$, $y^m=f^m(x^m)$, $y^t=f^t(x^t)$. The fusion model outputs a fused embedding from the video and text embeddings $y^{vt} = f^{vt}(y^v, y^t)$.

\vspace{0.5em}
\noindent \textbf{Fusion loss.} For training, we use an InfoNCE loss \cite{oord2018representation} between music and fused video-text embeddings: 
\begin{equation}
    \mathcal{L}_{vt\rightarrow m} = - \frac{1}{|\mathcal{D}|} \sum_{i \in \mathcal{D}} \frac{\exp(s(y^{vt}_i, y^m_i)/\tau)}{\sum_{j  \in \mathcal{D}} \exp{(s(y^{vt}_i, y^m_j)/\tau)}}
\end{equation}
where $s$ is a similarity function, $\mathcal{D}$ is 
a batch of data, and $\tau$ is a temperature hyperparameter we set as $\tau=0.03$.  For our similarity metric, we use the cosine similarity defined as $s(x,y) = x^T y / (\|x\| \cdot \|y\|)$. 
We note that the loss $\mathcal{L}_{vt\rightarrow m}$ is not symmetric as negatives are sampled from music embeddings only. So that our loss is symmetric, we instead train with the summed loss $\mathcal{L}_{m, vt} = \mathcal{L}_{vt \rightarrow m} + \mathcal{L}_{m\rightarrow  vt}$.

\vspace{0.5em}
\noindent \textbf{Text dropout.}
To address difficulties posed by the difference in granularity between audio/video and text, we introduce text dropout as a regularization mechanism.
With probability $p$, we set the input text embedding  $x^t$ to a specific value  $x^{\texttt{NULL}}$. In practice, we assign this $x^{\texttt{NULL}}$ input as the embedding produced by the pretrained $g^t$ model for an empty string. However, using a zero vector as $x^{\texttt{NULL}}$ works similarly.
In addition to improving the performance of music retrieval from text and video together, training with text dropout yields a model which also performs well at retrieval from video alone by removing dependence on text inputs.

\subsection{Implementation Details}

\noindent \textbf{Music tag generation.}  The key first step to our text generation process is obtaining the structured data describing each musical track. To do this, we use a music tagger trained on a dataset of music tracks manually annotated with a fixed pre-defined vocabulary of tags~\cite{lee2020metric}. 
Specifically,  the tagger predicts confidences for 41 instrument tags, 20 genre tags, and 28 mood tags. 
We aggregate these predictions at the clip or track level, and we filter the subsequent set based on confidence, keeping only those above a particular threshold (0.3 in our experiments).

\vspace{0.5em}
\noindent \textbf{ViML Model.} Following MVPt \cite{suris2022s}, we employ the Transformer architecture \cite{vaswani2017attention} for our music and video encoders $f^m$ and $f^v$. Transformers play a key role in improving model performance by encoding long-term context from video and music clips. 
We also use a similar two-layer Transformer architecture for our text encoder $f^t$ and the video-text fusion layer $f^{vt}$. However, we find that other fusion module architectures such as a single linear layer yield similar results. Please see our supplemental for study of fusion module architectures.

For base features, we use CLIP \cite{radford2021learning} to encode representations for video frames and text inputs, and DeepSim \cite{lee2020metric} to encode music. 
Following communication with the authors of \cite{suris2022s}, we split the video into 10-second segments and compute a feature for each segment by averaging CLIP embedding features computed at 6 frames per second. We compute 512 dimensional CLIP embedding features using OpenAI's CLIP ViT-B/32 model. 
We encode all input base features into embeddings of size $d=256$ using a linear projection layer for each modality. We also select $d=256$ as the output dimension for 
encoded video, text, music, and fused video-text representations from our model.

\newcommand{\width}{.18}

\begin{figure*}
    \captionsetup[subfigure]{labelformat=empty} %
    \centering
    \subfloat[A faint, simple acoustic piece of singing by a female vocalist with an acoustic guitar with a fast-paced strumming pattern in a closed room recorded live. great for singing along.] {
        \href{https://www.youtube.com/watch?v=sdKN2YzJxbY&t=89s}{   \includegraphics[width=\width\linewidth]{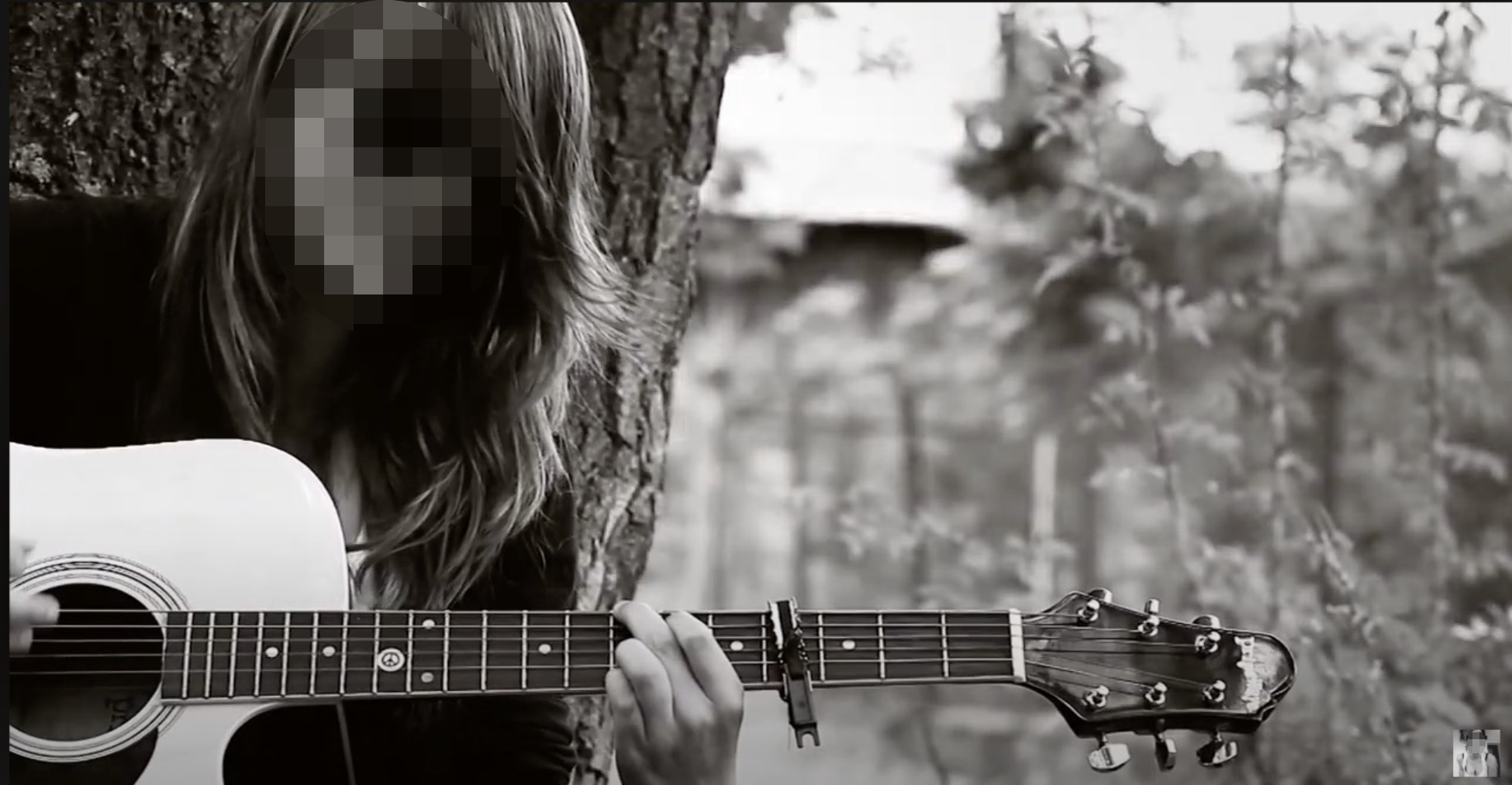}}%
    } \hspace{1mm}
    \subfloat[Hip-hop track with a dark synth pad with male aggressive rapping along with a chipmunk voice.]{
        \href{https://youtu.be/vAkBE67nc50?t=157}{   \includegraphics[width=\width\linewidth]{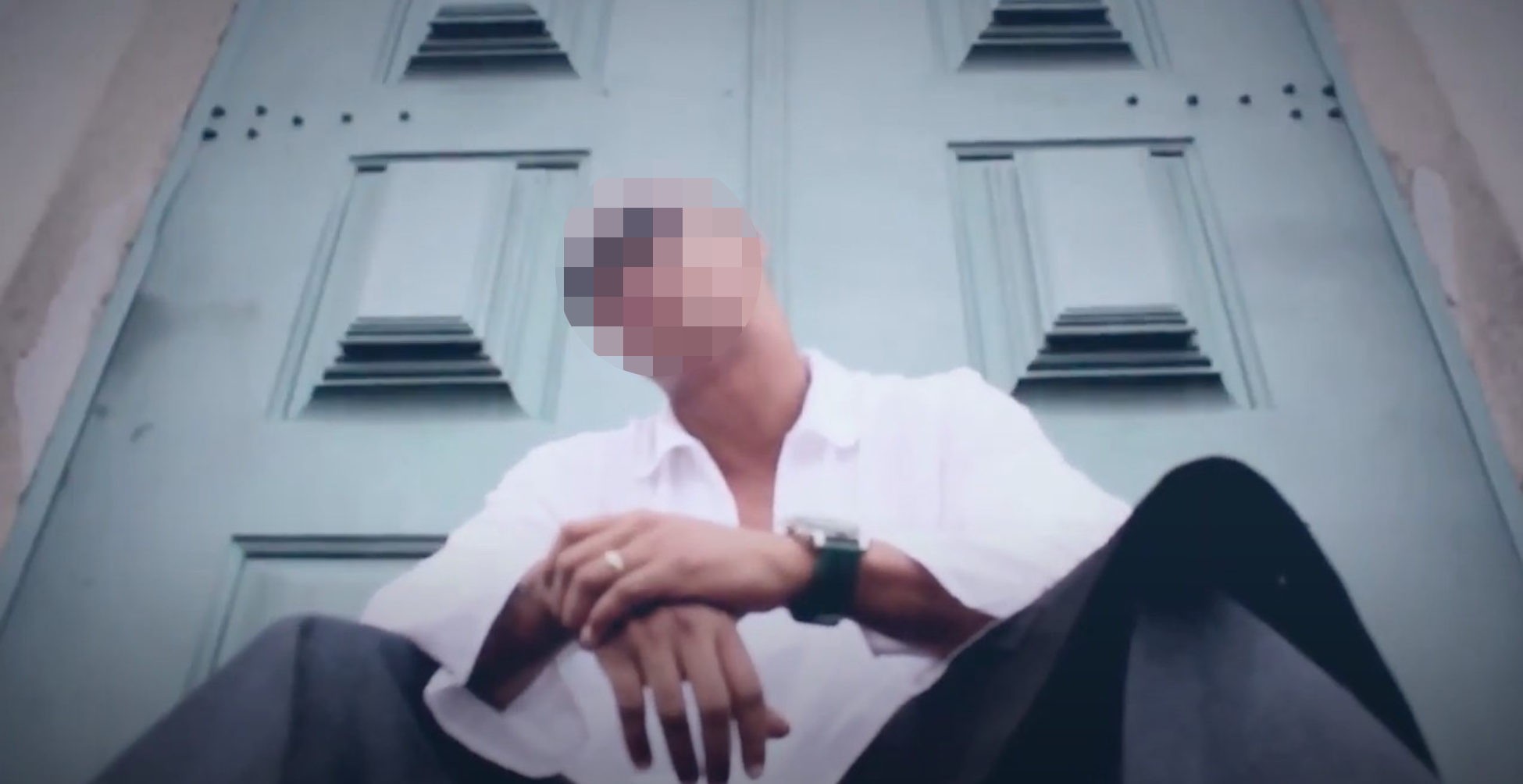}}%
    } \hspace{1mm}
    \subfloat[Instrumental track featuring an ambient pad and bell-like sounds. Seems to be a film score.]{
        \href{https://youtu.be/ze7p3GvtJDg?t=81}{   \includegraphics[width=\width\linewidth]{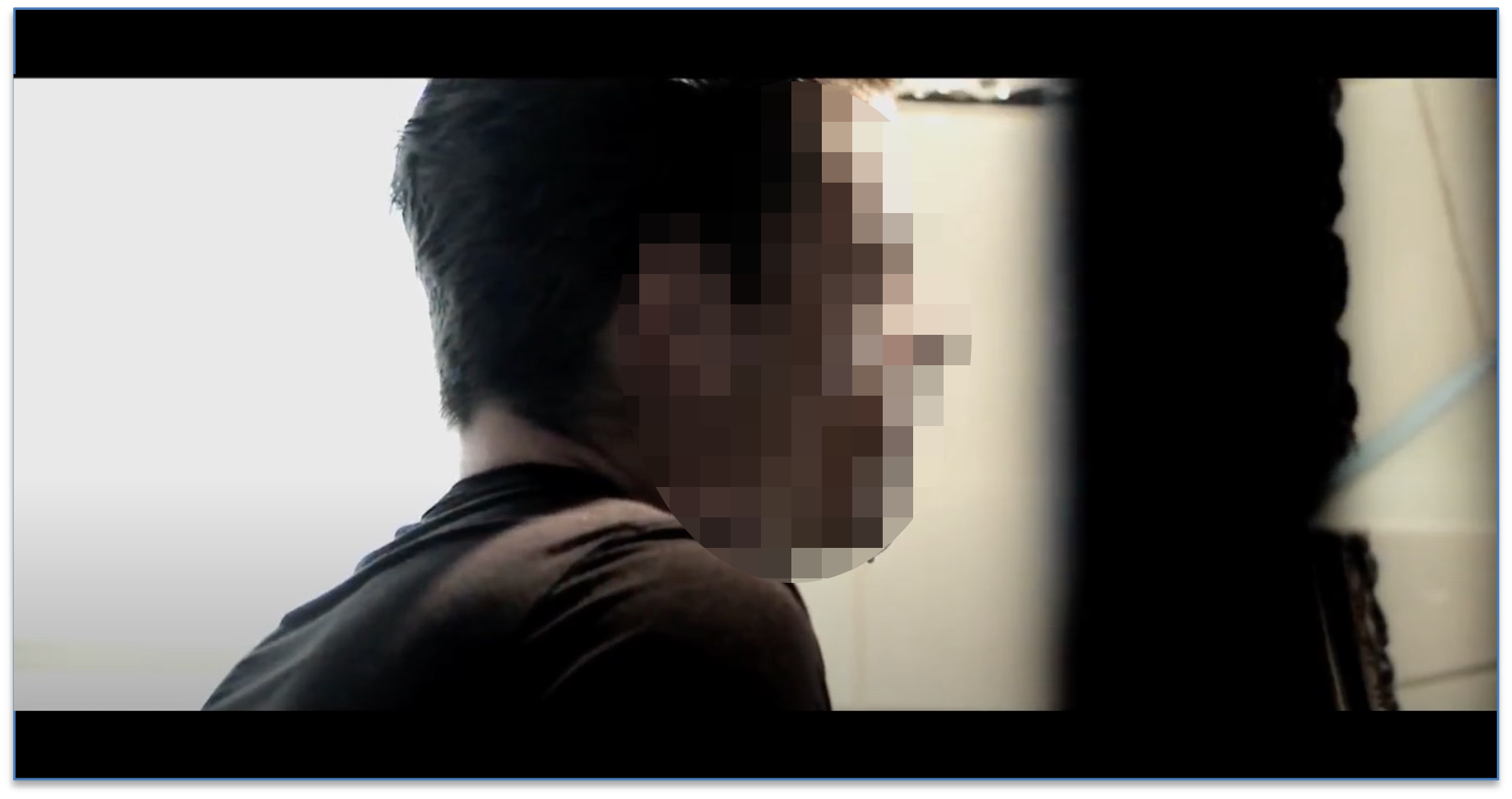}}%
    } \hspace{1mm}
    \subfloat[A cover version of Elvis Presley's pop song featuring male and female vocals and piano layers. Sounds like a confession in love.]{
        \href{https://www.youtube.com/watch?v=h5RvEeOfmmI&t=96}{   \includegraphics[width=\width\linewidth]{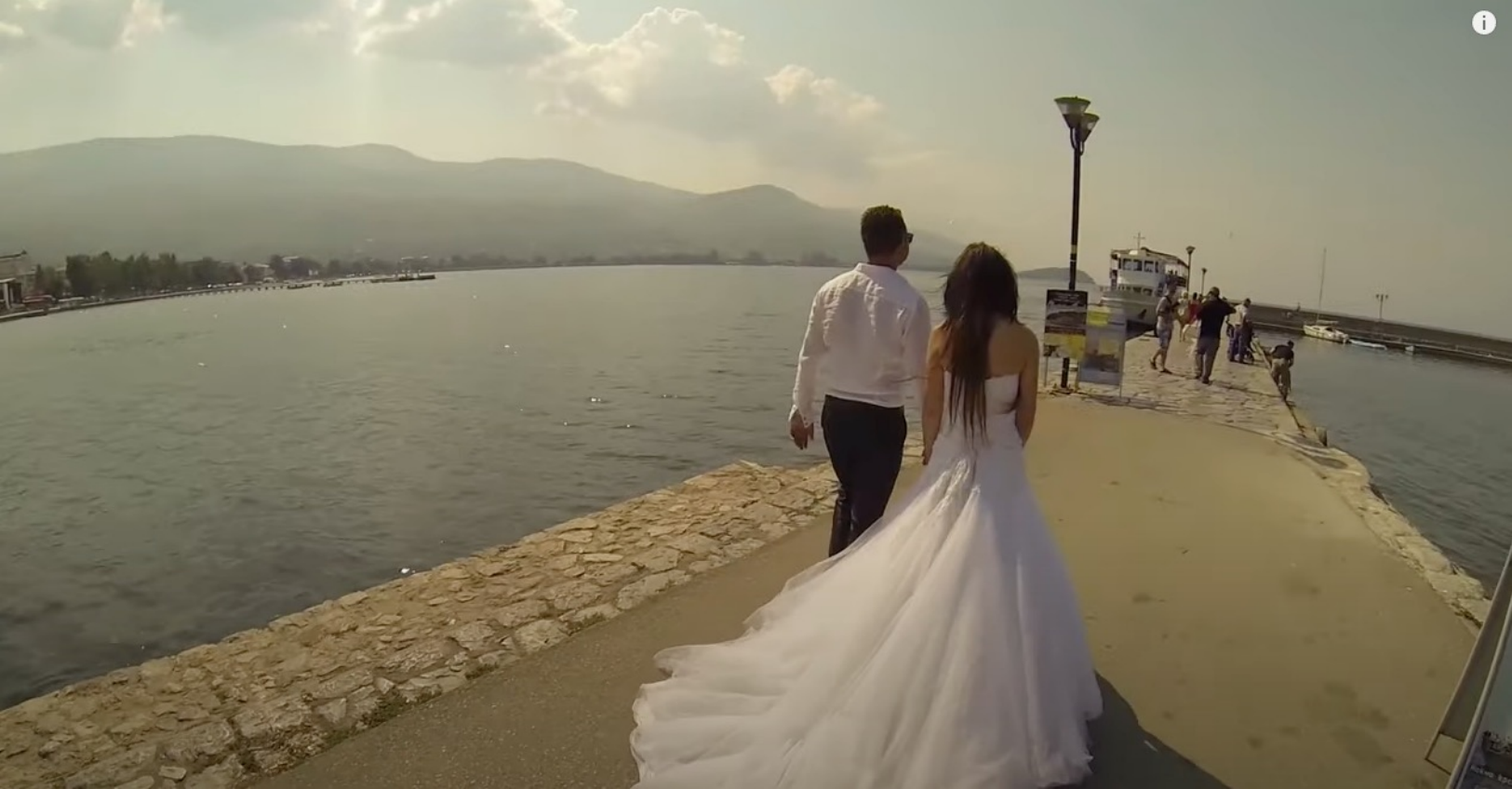}}%
    } \hspace{1mm} 
    \subfloat[Punjabi track with multiple vocals and other instruments that's great for traditional marriage celebrations.]{
         \href{https://www.youtube.com/watch?v=plcnjYyE6TE&t=95s}{   \includegraphics[width=\width\linewidth]{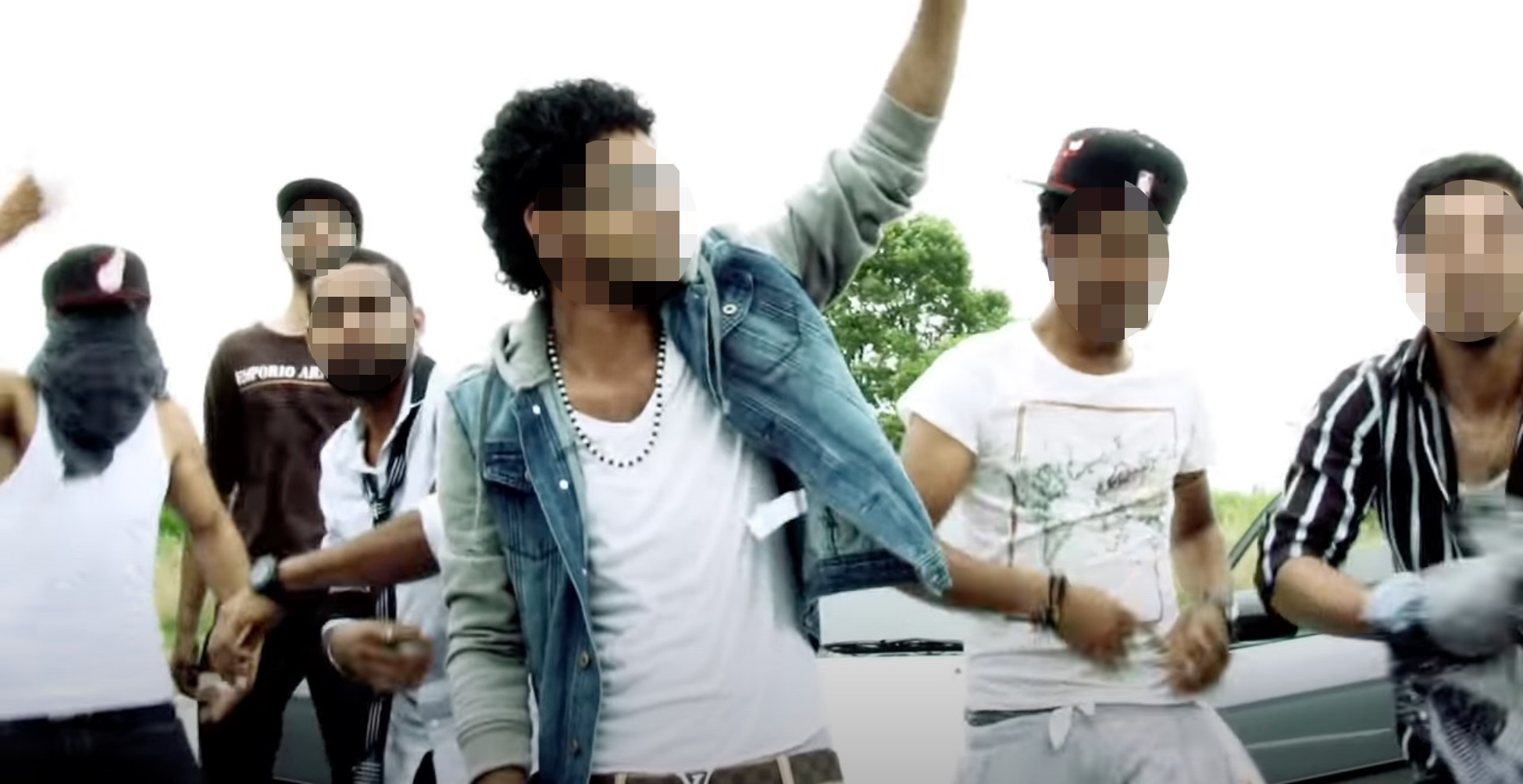}}%
    } \hspace{1mm}
    \caption{\textbf{Example annotations from our collected  YouTube8M-MusicTextClips dataset.} Each example shows a frame from the 10sec source video clip from which audio was extracted for annotation. Note that annotators were only provided \textit{audio} from the music video, so the annotation describes the music, but not the corresponding video. Each example in the figure contains a {\color{magenta} hyperlink} to the corresponding YT8M source video with timestamp at the start of the 10sec target clip. Hover over the video frame image and click to follow the link. 
    }
    \label{fig:yt8mtext_examples}
\end{figure*}

\section{Experiments}

In this section, we report our experimental settings and results. First, we describe our datasets and the evaluation protocol in \cref{sec:datasets}. Next, we investigate tag-based video-to-music retrieval, comparing against state-of-the-art video-to-music retrieval methods in \cref{sec:tag_exps}. In \cref{sec:lang_exps}, we evaluate performance of video-to-music retrieval guided by free-form text annotations. Finally, we perform ablation studies to measure the influence of text dropout in \cref{sec:dropout}.

\subsection{Datasets and Evaluation Protocol}
\label{sec:datasets}

\noindent \textbf{YT8M-MusicVideo.} In all of our experiments, we train models using the YT8M-MusicVideo dataset which includes around 100k videos with the ``music video" tag from the much larger YouTube8M dataset \cite{abu2016youtube}. We synthesize tags and a natural language text describing the music track of each video for the full dataset using the approaches described in \cref{sec:synthesis}. We also use the test split of YT8M-MusicVideo to evaluate tag-based retrieval in \cref{sec:tag_exps}. 

\vspace{0.5em}
\noindent  \textbf{YT8M-MusicTextClips.} 
In addition to the full YT8M-MusicVideo dataset, we also annotate a 4,000 sample subset of clips from YT8M-MusicVideo with human-provided text descriptions of the music track accompanying each video. 
To create these annotations, we sample 10 second audio clips from the middle of each music video, and we ask human annotators to describe the music they hear after listening to the audio clip. Thus, an annotation describes only the \textit{music} from a YT8M sample, and the annotators do not see the corresponding video. Example annotations are shown in Figure~\ref{fig:yt8mtext_examples} with links to the starting timestamp of the 10sec clips in corresponding YouTube videos. 
This annotated set is meant mainly for evaluation. As a result, the annotations are split into a larger set of 3,000 samples from the test set of YT8M-MusicVideo and a smaller set of 1,000 samples from the train set of YT8M-MusicVideo which we use as examples in the few-shot \texttt{prompt2text} synthesis process. We make the annotated text descriptions publicly available at our companion website\footnote{\url{https://www.danielbmckee.com/language-guided-music-for-video/index.html}}.

\vspace{0.5em}
\noindent  \textbf{Evaluation Set-up and Metrics.}
We evaluate music retrieval performance consistently with previous works \cite{pretet2021cross,suris2022s}.
However, in our case, a query can be either a video alone or a video and corresponding text annotation together.
For each query, we compute feature similarity between the query and a pool of N music tracks (we set N=2000 in the track-level setting and N=500 for evaluation on clips). The pool contains a single ground truth music track corresponding to the input query (the positive example) with the remaining music tracks in the pool being non-matching (\ie, negative examples).
We rank the music tracks in a query's pool by feature similarity, and find the rank of the query's ground truth matching music track (the positive example).
We then compute Recall@K (shortened to R@K) for K=1,5,10 and Median Rank, calculating the average of each of these metrics across the full set of test queries.

\begin{table*} 
\centering
\begin{tabular}{lcccccc}
\toprule
 Method                   & Train Text & Query Text Input & Median Rank $\downarrow$ & R@1 $\uparrow$ & R@5 $\uparrow$ & R@10 $\uparrow$ \\
\midrule
a. Pret\'et et al. \cite{pretet2021cross} & -          & -           & 234                      & 0.76           & 3.42           & 5.90            \\
b. MVPt \cite{suris2022s}                    &     -       &       -      & 13                       & 6.09           & 24.91          & 41.89           \\
c. MVPt+ \cite{suris2022s}                     & -                              & -                              & 5 & 27.93 & 50.64 & 60.68 \\
d. ViML (ours)              & \texttt{tags} & -                              & 3 & 29.43 & 62.49 & 75.40 \\
e. ViML (ours)                 & \texttt{tags} & \texttt{tags} & \textbf{2} & \textbf{49.49} & \textbf{81.61} & \textbf{89.41} \\
\midrule
f. Chance & & & 1000 & 0.05 & 0.25 & 0.50 \\
\bottomrule
\end{tabular}
\cprotect\caption{\textbf{Tag-based music retrieval on full YouTube8M-MusicVideo test set.} We compare ViML against prior methods on video to music retrieval without tag queries (row d.). We also evaluate ViML on video+text to music retrieval using (synthetic) tags at test time (row e.). The text descriptions for both training and evaluation are generated with the \texttt{tags} approach for these experiments.
}
\label{tab:track_results}
\end{table*}

\begin{table} 
\centering
\resizebox{0.95\linewidth}{!}{%
\begin{tabular}{llccccc}
\toprule
Method   & Train Text & MR $\downarrow$ & R@1 $\uparrow$ & R@5 $\uparrow$ & R@10 $\uparrow$ \\
\midrule
a. MVPt+ & -                                     & 17 & 12.20  & 29.43 & 40.46 \\
\midrule
b. ViML  &\texttt{tags}         & 15 & 11.95 & 30.34 & 42.62 \\
c. ViML  & \texttt{data2text} & 13 & 13.61 & 33.94 & 46.24 \\
d. ViML  & \texttt{prompt2text} & \textbf{12} & \textbf{14.09} & \textbf{35.04} & \textbf{47.88} \\
\midrule
Chance & & 250 & 0.20 & 1.00 & 2.00 \\
\bottomrule
\end{tabular}
}
\caption{
\textbf{Music retrieval with free-form natural language on YT8M-MusicTextClips test set.}  All methods which take text input are evaluated on the human text annotations as queries. Since the MVPt+ model does not take text inputs, it is evaluated on music retrieval from video alone for the same set of 3k video clips. MR is median rank.}
\label{tab:results}
\end{table}

\subsection{Tag-Based Retrieval}
\label{sec:tag_exps}

For our first set of experiments, we explore the setting of tag-based retrieval. Here the goal is to retrieve a music track given a query video together with a set of tags from a pre-defined vocabulary, such as ``happy", ``piano" and ``jazz".  This setting could be practically interesting in some applications, \eg, tag-based search. To address this setting, we train our model on text synthesized with the \texttt{tags} approach.
In these experiments, we train a track-level model and perform retrieval on a
track-level in a manner consistent with prior work \cite{pretet2021cross,suris2022s}. To directly compare results with prior work, we perform retrieval on the full YT8M test set consisting of around 10K samples. As shown in Table \ref{tab:track_results}, we include three baselines: the model proposed by Pret\'et et al.~\cite{pretet2021cross}, the MVPt model~\cite{suris2022s}, and an improved version of MVPt that we call MVPt+, where 
we tune the temperature parameter $\tau$ in the InfoNCE loss to 0.03. This change leads to further significant improvement in performance.

Next, we introduce our model (ViML) trained on data generated from the \texttt{tags} approach. We evaluate our ViML model in two settings. First, we evaluate without input texts at test time (an empty text input is used instead). Second, we evaluate with text inputs at test time. As we do not have track-level human-provided music tag annotations for the full YT8M-MusicVideo split, we evaluate the track-level model on synthetically generated tags using our \texttt{tags} approach. 
While a model trained on the \texttt{tags} synthesized data may not generalize to out-of-domain free-form text inputs, the tag-based prompting can be a convenient way to guide music retrieval with key desired attributes (for example ``female vocalist, guitar, happy").
The tag-based retrieval we report can serve as an upper bound for this type of user tag-guided retrieval since the tag-based text for testing comes from the same music tagger model we used to synthesize training data.

Evaluating our model with synthetic tags leads to a very substantial performance increase over MVPt+ of 20-30 points in each recall metric. Interestingly, our ViML model evaluated without text at test time not only matches the video-to-music retrieval performance of MVT+ but substantially improves over MVPt+, especially in Recall@5 and Recall@10. 
This performance increase is not simply a result of added parameters in the fusion layer, as a fusion module consisting of only a single linear layer yields similar results (see our supplemental for further details). 
This result suggests that training jointly with the text domain can lead to improvements in the video and audio representations. We hypothesize that the joint training with language helps to disentangle the video-audio space into semantically meaningful dimensions corresponding to the provided tags as well as helps to suppress non-relevant dimensions, \eg, corresponding to presence/absence of some non-relevant objects.

\begin{table*} 
\centering
\begin{tabular}{llcccccc}
\toprule
Method   & Train Text & Dropout & Text Inputs & Median Rank $\downarrow$ & R@1 $\uparrow$ & R@5 $\uparrow$ & R@10 $\uparrow$ \\
\midrule
a. MVPt+    &                -                       &  -  &             -            & 17         & 12.20 & 29.43 & 40.46 \\
\midrule
b. ViML & \texttt{prompt2text} & \xmark & - & 20 & 9.94  & 26.42 & 37.01  \\
c. ViML & \texttt{prompt2text} & \xmark & \texttt{human}                               & 15 & 11.45 & 30.45 & 42.77  \\
d. ViML & \texttt{prompt2text} & \cmark & - & 16 & 12.27 & 30.34 & 41.51  \\
e. ViML & \texttt{prompt2text} & \cmark & \texttt{human}                               & \textbf{12} & \textbf{14.09} & \textbf{35.04} & \textbf{47.88}  \\
\bottomrule
\end{tabular}
\caption{
\textbf{Influence of training with text dropout on retrieval performance.} Evaluated on the YT8M-MusicTextClips test set. }
\label{tab:dropout}
\end{table*}

\subsection{Free-Form Natural Language Retrieval}
\label{sec:lang_exps}

For the next experiments, we turn to retrieval with free-form natural language inputs. The goal is, given an input video and a query free-form natural language description, to retrieve a relevant music track.
For this setting, we evaluate on testing videos from the YT8M-MusicTextClips dataset which contains free-form human text annotations describing the music corresponding to each video in the dataset.

In these experiments, we use a similar protocol to the ``segment-level" setting reported by Sur\'is et al. \cite{suris2022s}, but our input video includes only a 30sec clip surrounding the 10sec of audio labeled by a human annotator. In contrast, a model had access to a large context spanning the full source video in the previous segment-level setting reported by Sur\'is et al. \cite{suris2022s}. We note that retrieval in this setting is significantly more difficult than the segment-level setting in \cite{suris2022s} or the track-level setting reported in \ref{sec:tag_exps} due to the limited context. However, such retrieval is of particular interest given the rise of short-form video in social media and entertainment.

Results are summarized in Table \ref{tab:results}. Our baseline is an MVPt+ model which has been trained on 30sec segments (training MVPt+ on full videos and testing on 30sec clips causes a much more severe drop in performance). %
We next report music retrieval using video and free-form human text descriptions as input queries to our ViML model.
In Table \ref{tab:results}, we report three variants trained on YT8M music videos with text synthesized by each of the three approaches described in Sec. \ref{sec:synthesis}. The model trained with our first \texttt{tags} synthesis baseline (b.)~provides substantial improvement over retrieval with MVPt+ using only video (a.). Next, we evaluate the \texttt{ data2text} approach (c.) which generates more natural phrases while strictly preserving tag semantics. This approach provides a consistent improvement over the ViML \texttt{tags} variant (b.). Finally, our \texttt{prompt2text} approach (d.)~leads to the best performance showing that large language models prove to be strong annotators on this task with careful few-shot prompting.

\begin{figure}
  \begin{center}
  \includegraphics[width=0.48\textwidth]{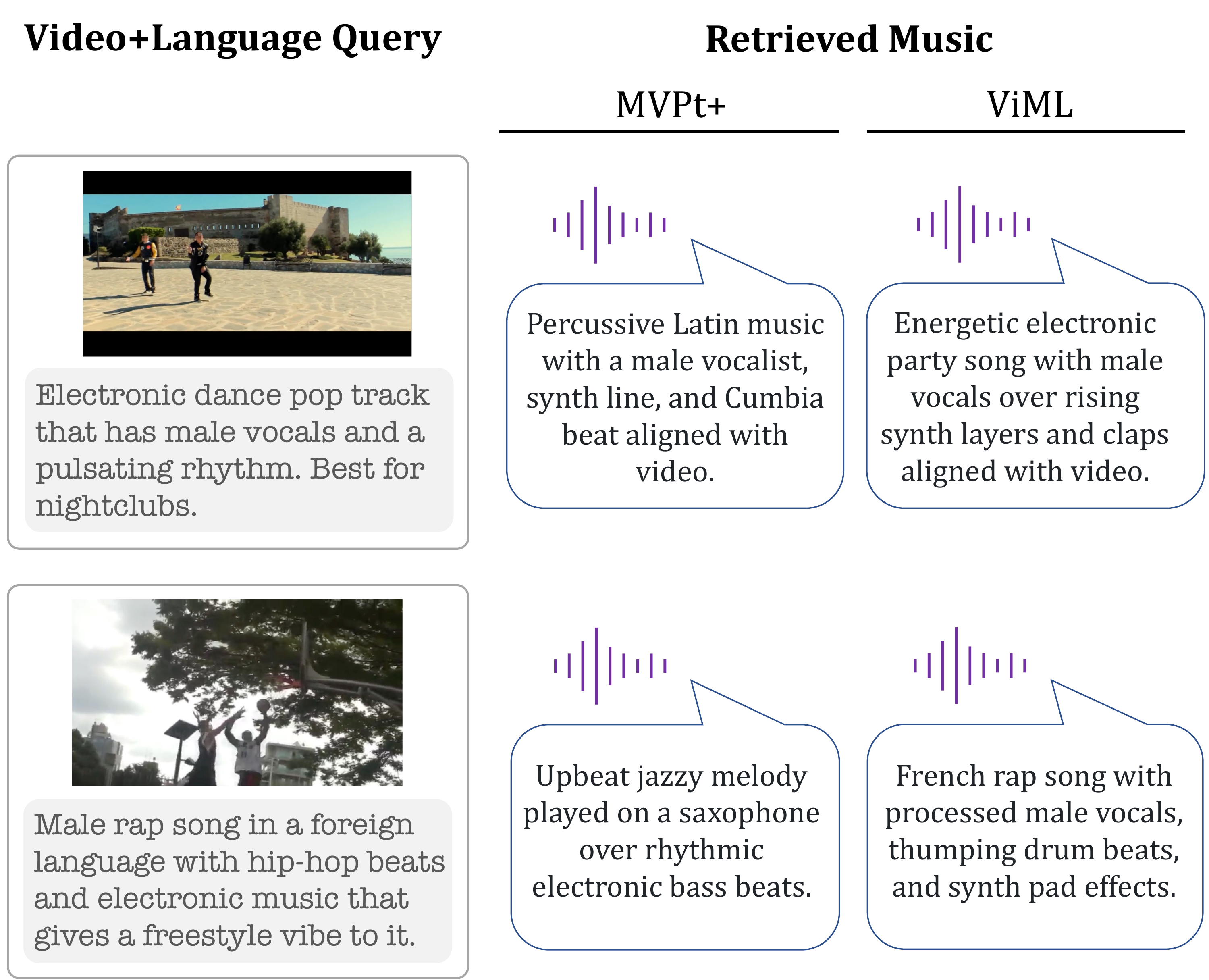}
  \caption{
  \textbf{Qualitative results on YouTube8M-MusicTextClips test set.} We compare music retrieval quality for two examples using the MVPt+ model and our ViML model.
  The column on the left includes a frame from the input video and the input text description describing the target music. 
  The MVPt+ model takes only the video as an input while the ViML model takes both video and corresponding text. The two columns on the right contain retrieved music for MVPt+ and ViML respectively. 
  \textbf{Please see results in the companion video on our website.}
  }
  \label{fig:qualitative}
  \end{center}
  \end{figure}

\noindent \textbf{Qualitative results.} In Figure \ref{fig:qualitative}, we provide qualitative retrieval results for examples in YouTube8M-MusicTextClips. In the first example, both models retrieve tracks that match the style and beat of the input video well. However, only the ViML can match the correct musical style by using the input text. In the second example, only the ViML result correctly matches the desired music genre and the mood of the video.

\subsection{Analysis of Text Dropout}
\label{sec:dropout}

In Table \ref{tab:dropout}, we compare the performance of our ViML model trained on \texttt{prompt2text} descriptions with and without text dropout. 
We evaluate this model on music retrieval in two settings: (i) using only video (inputting empty text, rows b.\ and d.) as a query and (ii) using  both video and human text descriptions together as a query (rows c.\ and e.). 
As expected, adding text dropout during training (d.) improves the performance of retrieval using only video (b.). However, interestingly, text dropout also substantially improves performance when the query includes natural language  (e.\ vs.\ c.), suggesting that text dropout is a very useful regularization technique in the multimodal setting. 
We find that without text dropout, training begins to plateau early as the model starts overfitting to the training text inputs.
Since video is a much richer and more complex modality, forcing more attention to this modality during training improves learning.
We find that the dropout technique is most effective at high rates of dropout in the range 0.8-0.95, and we use a dropout rate of 0.8 in all of our experiments.

\section{Conclusion}

In this work, we introduced an approach to allow language-guided music recommendation for video.
We proposed a model, ViML, which fuses text and video inputs to find music matching both domains and introduced the text dropout technique to improve training.
To obtain data for training, we proposed a free-form music description synthesis approach using a large language model (BLOOM-176B) and outputs from a pretrained music tagger. 
Our results show that large language models provide a powerful tool for training data synthesis in domains where text data is limited but other structured data is available.
To evaluate our method, we also introduced a new dataset, YouTube8M-MusicTextClips, which includes high quality free-form human descriptions of the music in YT8M videos.
There are many exciting directions to build upon this work including allowing more fine-grained control over specific music attributes or language-guided audio-video generation.

\clearpage

{\small
\bibliographystyle{ieee_fullname}
\bibliography{egbib}
}

\clearpage

\appendix
\setcounter{page}{1}

\twocolumn[
\centering
\Large
\textbf{Appendix} \\
\vspace{1.0em}
] 
\appendix

\renewcommand\thesection{\Alph{section}}



\section{Fusion Module Architecture Study}
\label{sec:fusion_arch}

\begin{table*} 
\centering
\begin{tabular}{lcccccc}
\toprule
Method   & \# Parameters & Median Rank $\downarrow$ & R@1 $\uparrow$ & R@5 $\uparrow$ & R@10 $\uparrow$ \\
\midrule
a. Addition              & 0     & 12 & 13.73 & 34.12 & 46.52 \\
b. Linear                & 131K & 13 & 13.19 & 33.45 & 45.68 \\
c. MLP                   & 1.6M  & 13 & 13.20 & 32.69 & 44.94 \\
d. Transformer (1 layer) & 1.4M  & 12 & 13.98 & 34.95 & 46.85 \\
e. Transformer (2 layer) & 2.8M  & 12 & \textbf{14.09} & \textbf{35.04} & \textbf{47.88} \\
\bottomrule
\end{tabular}
\caption{
\textbf{Study of fusion layer architecture.} All models are trained on the synthesized \texttt{prompt2text} data, and we report results on the YT8M-MusicTextClips 3k test set.}
\label{tab:fusion_module}
\end{table*}

We evaluate different architectures for the fusion module that is responsible for combining encoded visual and text inputs. In Table \ref{tab:fusion_module}, we present five architecture variants. First, we benchmark fusion by direct addition of the visual and text Transformer encoder outputs (a.) which removes learned parameters from the fusion module entirely. Next, we evaluate three different learned fusion module architectures which involve passing the concatenated visual and text features as input to: (b.) a single linear layer (c.) a two-layer MLP, (d.) a 1-layer Transformer network, and (e.) a 2-layer Transformer network. 
We find that the size of the fusion module does not significantly change performance. 
We use the 2-layer Transformer fusion architecture in our main results given the slightly higher performance in recall metrics, but similar performance can be achieved with the other fusion architectures including the ``addition" fusion module which does not include learned parameters.

\section{Training without Video}\label{sec:musictext}

We also experimented with models trained only on music and text but found these models to significantly underperform other baselines at music retrieval on the YT8M-MusicTextClips test set. This is not surprising as the input video contains a great deal more information than the short human text descriptions in the dataset. Performance of our music+text model trained on \texttt{prompt2text} data and evaluated on human texts in the YT8M-MusicTextClips test set was
 Recall@1/5/10=2.52/9.27/15.52 and MR=56
 (compare to Table {\color{red} 2} results from the main paper).
We report the results of our track-level music+text model trained with tag inputs as MT in Table \ref{tab:track_musictext_ens_results} (a.) (compare to Table {\color{red} 1} results from the main paper). This model also performed substantially below MVPt+ or ViML.

\begin{table*} 
\centering
\begin{tabular}{lcccccc}
\toprule
 Method                   & Train Text & Query Text Input & Median Rank $\downarrow$ & R@1 $\uparrow$ & R@5 $\uparrow$ & R@10 $\uparrow$ \\
\midrule
a. MT     &   \texttt{tags} & \texttt{tags}        &  15 & 11.51 & 30.36 & 42.72  \\
b. MVPt+ [{\color{green} 37}]                     & -                              & -                              & 5 & 27.93 & 50.64 & 60.68 \\
c. ViML (ours)                 & \texttt{tags} & \texttt{tags} & 2 & 49.49 & 81.61 & 89.41 \\
\midrule
d. MT \& MVPt+ Ens.   &  \texttt{tags} & \texttt{tags} & 1  & 55.95 & 81.73 & 88.82  \\
e. ViML \& MVPt+ Ens. &  \texttt{tags} & \texttt{tags}  & 1  & 59.86 & 85.14 & 91.43  \\
f. ViML \& MT  Ens.    & \texttt{tags} & \texttt{tags}  & 1  & \textbf{63.05} & \textbf{91.32} & \textbf{96.59} \\
\midrule
g. Chance & & & 1000 & 0.05 & 0.25 & 0.50 \\
\bottomrule
\end{tabular}
\cprotect\caption{\textbf{Tag-based music retrieval on full YouTube8M-MusicVideo test set for music+text model and model ensembles.} 
For convenient comparison, we also report the MVPt+ and ViML results from Table {\color{red} 2} in the main text.
We denote the music+text model described in Sec. \ref{sec:musictext} as MT in the table.
}
\label{tab:track_musictext_ens_results}
\end{table*}

\section{Ensembled Models}

In addition to the baselines reported in Table {\color{red} 1} of the main text, we also investigated forming a stronger baseline by combining MVPt+ and the music+text model from ~\cref{sec:musictext} into an ensemble. 

More specifically, for a music track $m$ and a corresponding video, text pair $(v, t)$, we compute the total similarity score as a weighted sum $(1 - \alpha) \cdot s(y^v, y^m) + \alpha \cdot s(z^t, z^m)$ where $y^v, y^m$ are the video and music embeddings generated by MVPt+, $z^t, z^m$ are the text and music embeddings generated by our music+text model, and $\alpha$ is a coefficient which we tuned. 

As shown in Table \ref{tab:track_musictext_ens_results} (d.), we found this music+text model and MVPt+ ensemble to reach strong performance, exceeding Recall@1 performance of ViML and achieving similar Recall@5/10 ViML performance. However, we found that such ensembling could be used to improve the performance of ViML as well. In particular, computing scores for music retrieval as a weighted sum of similarity scores from ViML and MVPt+ led to substantial improvements over ViML performance as shown in Table \ref{tab:track_musictext_ens_results} (e.). An ensemble of ViML and our music+text model led to the highest performance in Table \ref{tab:track_musictext_ens_results} (f.).

\section{Music Matching the Pace of Videos}

In our qualitative results, we did not observe many examples where the music beats per minute (BPM) does not match the video pace. We hypothesize that a given music genre lives in a limited tempo range. Therefore, being able to match effectively the music genre may return a well-matching tempo for free. Note that we do not have fine-grained tempo alignment, \eg, depicted dance motions may not be perfectly in sync with the music. One possible future direction could be to refine the alignment between the music and depicted action in the video.

\section{Text Synthesis Examples Outputs}
\label{sec:text_examples}
\begin{figure*}[h!]
  \begin{center}
  \includegraphics[width=1.0\textwidth]{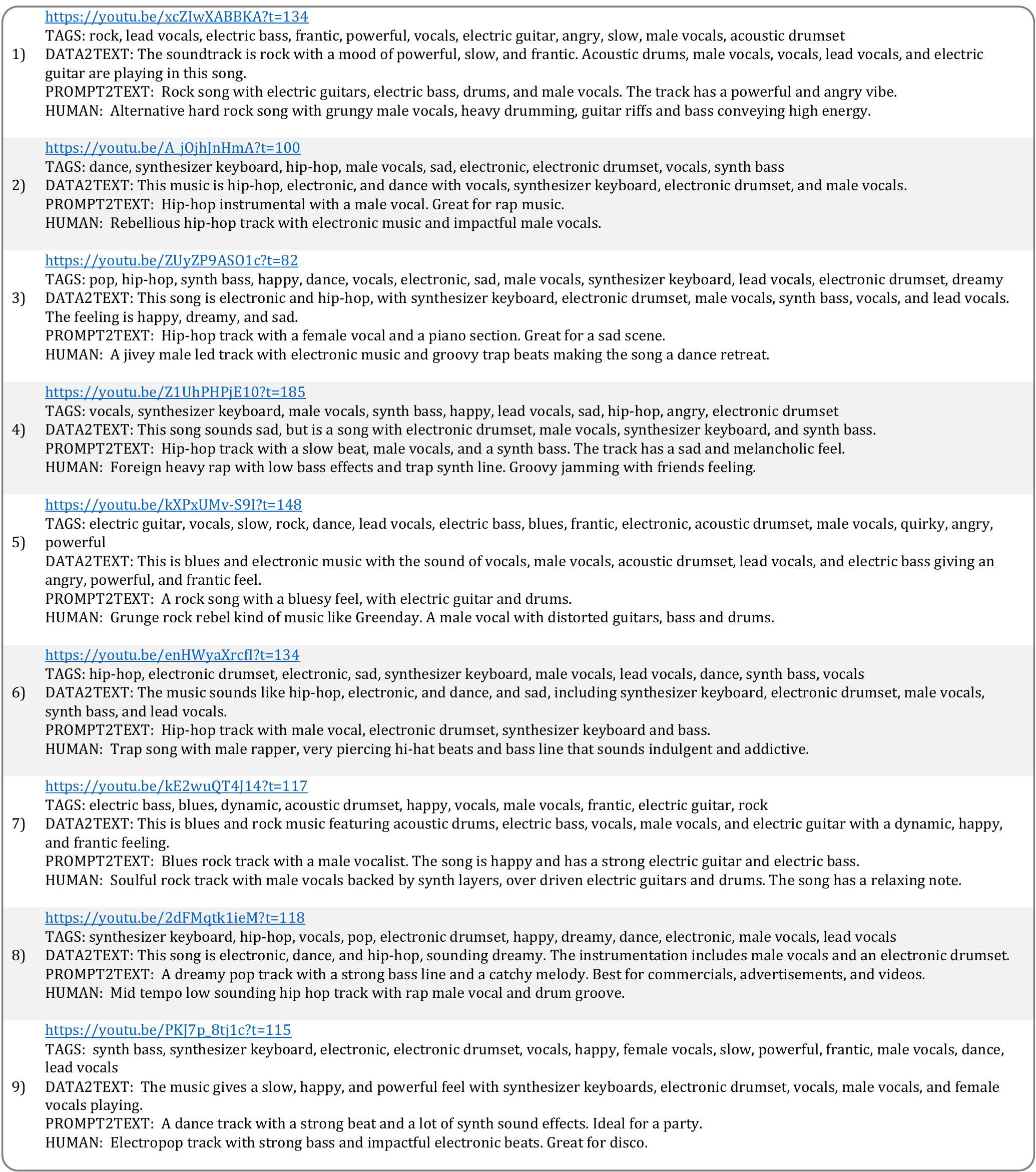}
  \caption{\textbf{Synthesized text examples using our approach}. We randomly select examples from the YouTube8M-MusicTextClips test set. We show the output from our \texttt{tags}, \texttt{data2text}, and \texttt{prompt2text} approaches for each example video. We also show real \texttt{human} annotations collected for each example. 
  The text synthesis approaches show tradeoffs between preserving tag accuracy and increasing diversity of vocabulary and phrase structure. The \texttt{prompt2text} approach generates outputs with the highest diversity and most closely resembling human annotations.
    Additional examples shown in Figure \ref{fig:text_synthesis_examples_pg2}.
  }
  \label{fig:text_synthesis_examples}
  \end{center}
  \end{figure*}

\begin{figure*}[h!]
  \begin{center}
  \includegraphics[width=1.0\textwidth]{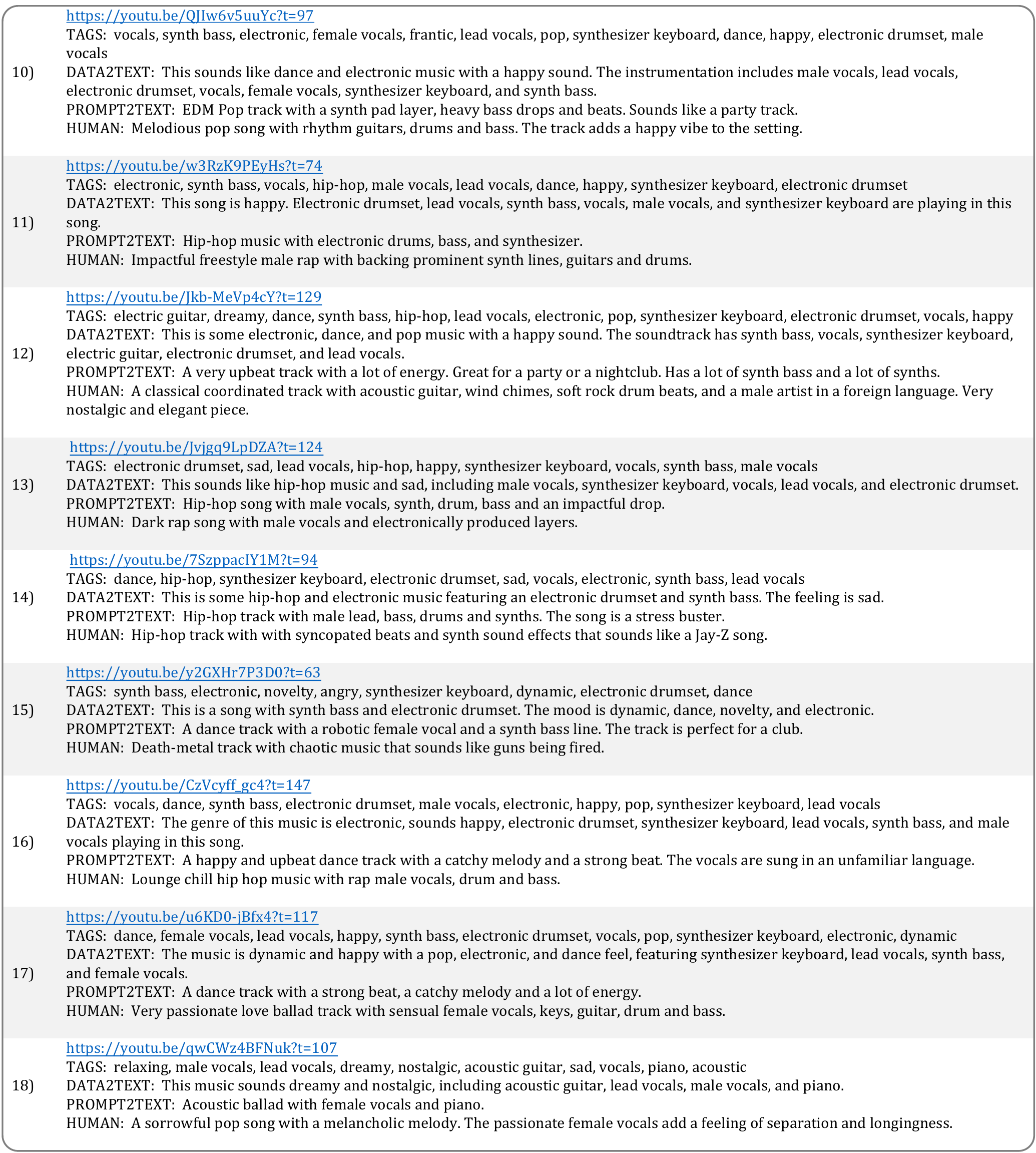}
  \caption{\textbf{Synthesized text examples using our approach}. Continued from Figure \ref{fig:text_synthesis_examples}.
  }
  \label{fig:text_synthesis_examples_pg2}
  \end{center}
  \end{figure*}

In \Cref{fig:text_synthesis_examples,fig:text_synthesis_examples_pg2}, we present generated outputs from our text synthesis approaches along with real human annotations for randomly selected examples from the YouTube8M-MusicVideo dataset.
The text synthesis approaches show different tradeoffs between tag accuracy and diversity of form/language. 

The \texttt{prompt2text} setting is the most free-form text synthesis approach but will sometimes generate outputs which are not true to the original tag predictions for a track. In general, the \texttt{prompt2text} descriptions tend to be shorter and often omit information in the input tags. The language model used in the \texttt{prompt2text} approach can also sometimes hallucinate information which is completely wrong (e.g. ``female vocal and a piano section" in \cref{fig:text_synthesis_examples} ex.\ 3). However, the diversity of vocabulary and structure in the outputs produced by \texttt{prompt2text} makes this approach most similar to real human annotations.

\end{document}